\documentclass[lettersize,journal]{IEEEtran}
\usepackage{amsmath,amsfonts}
\usepackage{algorithmic}
\usepackage{algorithm}
\usepackage{array}
\usepackage[caption=false,font=normalsize,labelfont=sf,textfont=sf]{subfig}
\usepackage{textcomp}
\usepackage{stfloats}
\usepackage{url}
\usepackage{verbatim}
\usepackage{graphicx}
\usepackage{array}
\usepackage{bm}
\usepackage{makecell}
\usepackage{diagbox}
\usepackage{booktabs}
\usepackage{graphicx}
\usepackage{subfig}
\usepackage{graphicx}
\usepackage{bbding}
\usepackage {gensymb}
\usepackage{algorithm} 
\usepackage{multirow}
\usepackage[T1]{fontenc}

\usepackage[export]{adjustbox}
\usepackage{multicol}

\makeatletter

\newcommand{\Rmnum}[1]{\expandafter\@slowromancap\romannumeral #1@}
\makeatother

\usepackage{stfloats}
\usepackage{longtable}
\usepackage{xcolor}

\usepackage[nocompress]{cite}

\usepackage[
pdfauthor={derajan},
pdftitle={How to do this},
pdfstartview=XYZ,
bookmarks=true,
colorlinks=true,
linkcolor=blue,
urlcolor=blue,
citecolor=blue,
pdftex,
bookmarks=true,
linktocpage=true,   
hyperindex=true
]{hyperref}

\begin{document}
	
	\title{Unlocking Optical Prior: Spectrum-Guided Knowledge Transfer for SAR Generalized Category Discovery}
	
	\author{Jingyuan Xia, Ruikang Hu, Ye Li, Zhixiong Yang, Xu Lan, Zhejun Lu
		\thanks{
			This work was supported in part by the National Natural Science Foundation of China under Projects 62576350, 62406330, and 625B2180. (Corresponding author: Zhejun Lu)
			
			Jingyuan Xia, Ruikang Hu, Ye Li, Zhixiong Yang, and Zhejun Lu are with the College of Electronic Science and Technology, National University of Defense Technology, Changsha, 410073, China (e-mail: j.xia10@nudt.edu.cn; huruikang@nudt.edu.cn; liye19@nudt.edu.cn; yzx21@nudt.edu.cn; luzhejun@nudt.edu.cn)
			
			Xu Lan is with the State Key Laboratory of Complex System Simulation and Modeling Technology, Beijing, 100091, China (e-mail: lanxu1991@yeah.net)
		}
	}
	
	\maketitle
	
	\maketitle
	
	\begin{abstract}
		Generalized Category Discovery (GCD) holds significant promise for the label-scarce Synthetic Aperture Radar (SAR) domain, yet its efficacy is severely constrained by the cross-modal incompatibility between the inherent optical prior of the Large Vision Models (LVMs) and SAR imagery. Existing domain adaptation methods often lack an inductive bias that reflects imaging characteristics, consequently failing to effectively transfer optical prior into the SAR domain. To address this issue, the Modal Discrepancy Curve (MDC) is introduced to model cross-modal discrepancy as a structured frequency-domain descriptor derived from spectral energy distributions. Leveraging this formulation, we propose the MDC-guided Cross-modal Prior Transfer (MCPT) framework, a pre-training paradigm that operates on paired optical–SAR data. Within this framework, Adaptive Frequency Tokenization (AFT) converts the MDC into learnable tokens, and Frequency-aware Expert Refinement (FER) performs band-wise discrepancy-aware feature refinement using these tokens. Based on the refined representations, contrastive learning aligns refined embeddings across modalities and internalizes the adaptation pattern. Ultimately, the superior SAR feature representation capability learned during paired pre-training is applied to downstream single-modal SAR-GCD tasks. Extensive experiments demonstrate state-of-the-art performance across multiple mainstream datasets, indicating that frequency-domain discrepancy modeling enables more effective adaptation of optical prior to SAR imagery.
	\end{abstract}

	
	\section{Introduction}
	\IEEEPARstart{S}{ynthetic} Aperture Radar (SAR), with its all-weather and day-and-night imaging capability, plays an indispensable role in remote sensing applications such as environmental monitoring and disaster warning \cite{1_0_1,1_0_2,1_0_5,1_0_6}. However, the interpretation of SAR imagery heavily relies on expert knowledge, which inevitably incurs high annotation costs and long processing cycles \cite{1_0_3,1_0_12}. As a result, fully annotated SAR datasets are often unavailable in real-world scenarios, leading to a practical data regime where a small amount of labeled data coexists with a large amount of unlabeled data, further complicating target recognition \cite{1_0_4, 1_0_11}. In this context, Generalized Category Discovery (GCD) provides a promising solution for SAR understanding, as it aims to recognize known classes while simultaneously discovering previously unseen categories from unlabeled samples \cite{2_0_2, 1_0_10}. Compared with optical imagery, SAR data are governed by more complex imaging and scattering characteristics \cite{1_0_7,1_0_8,1_0_9}, making intra-class variation and inter-class discrimination more challenging. Consequently, SAR-GCD places a stronger demand on the quality and robustness of representation prior.
	
	Current GCD methods can generally be divided into two streams: non-parametric classifier approaches \cite{1_1_1,1_1_2,1_1_3,1_1_4,1_1_5} and parametric classifier approaches \cite{1_2_1,1_2_2,1_2_3,1_2_4,1_2_5,1_2_6}. Non-parametric approaches typically organize the feature space through contrastive learning and clustering. However, the optimization of the encoder is inherently decoupled from the final classification process, and in the absence of reliable clustering guidance, self-supervised learning on unlabeled data may even separate latent intra-class samples in the feature space. Parametric approaches, in contrast, jointly optimize the encoder and classifier by assigning pseudo-labels to unlabeled data and selecting high-confidence instances for supervision, and have therefore become the mainstream paradigm. Nevertheless, their effectiveness heavily depends on pseudo-label quality. Once noisy pseudo-labels are introduced, confirmation bias can accumulate and substantially impair novel-class recognition. In essence, both streams rely critically on the quality of the learned representations \cite{1_3_1,1_3_2}. This dependence becomes even more pronounced in the SAR domain, where complex scattering characteristics and limited annotations make robust representation learning particularly challenging. In practice, such representations are increasingly inherited from large-scale pretrained vision models. Therefore, the key challenge in SAR-GCD lies not only in designing more reliable clustering or pseudo-labeling strategies, but in effectively adapting strong optical prior to the heterogeneous SAR feature space.
	
	In general, the success of GCD is closely tied to the powerful semantic prior provided by pretrained Large Vision Models (LVMs). To verify this point, we first re-evaluate representative GCD frameworks with the advanced DINOv2 and establish a DINOv2-based benchmark for SAR-GCD. The results show that stronger optical prior can indeed raise the recognition baseline substantially, yet their straightforward transfer to the SAR domain still suffers from clear performance limitations. The root cause stems from the intrinsic imaging discrepancy between optical and SAR modalities, rather than from the insufficiency of clustering or pseudo-labeling strategies, which prevents optical prior from being directly adapted to the heterogeneous SAR feature space. As a result, clustering guidance in non-parametric methods becomes less reliable, while pseudo-labels in parametric methods are more easily corrupted. Meanwhile, most existing domain adaptation methods mainly rely on implicit end-to-end feature alignment, treating modality discrepancy as a black-box optimization target without explicitly modeling underlying object characteristics \cite{1_3_3,1_3_4,1_3_5,1_3_8}. This deficiency limits the stable transfer of optical prior from LVMs \cite{1_3_6,1_3_7} into the SAR domain.
	
	To address the above issue, we make a key empirical observation that although the discrepancy between optical and SAR imagery is highly entangled in the pixel domain, it exhibits much more structured energy distributions in the frequency domain. This observation suggests that modality discrepancy can be explicitly characterized in spectral space, rather than being handled solely through implicit feature alignment. Motivated by this property, we introduce the Modal Discrepancy Curve (MDC), which describes the spectral energy distribution of optical-SAR residuals in the frequency domain. The MDC provides both a visualizable description of cross-modal discrepancy and a quantifiable frequency-domain representation for subsequent feature alignment.
	
	Building upon the MDC, we further propose the MDC-guided Cross-modal Prior Transfer (MCPT) pre-training strategy to efficiently adapt optical prior to the SAR domain. Specifically, MCPT first employs Adaptive Frequency Tokenization (AFT) to convert the MDC into spectral tokens that encode frequency-domain discrepancy patterns, thus transforming frequency-domain discrepancy into learnable guidance signals. These tokens are then fed into a Frequency-aware Expert Refinement (FER) module, which performs band-wise discrepancy-aware feature refinement. Finally, contrastive learning is used to enforce cross-modal consistency of the refined embeddings, while back-propagation progressively internalizes the alignment capability into the shared backbone parameters. Through this dual mechanism of spectral guidance and parameter adaptation, the optical prior embedded in DINOv2 is more effectively adapted to SAR representations, thus benefiting downstream SAR-GCD tasks. Moreover, to disentangle this improvement from the effect of paired cross-modal pre-training itself, we further introduce a paired pre-training baseline without MDC guidance.
	
	The main contributions of this work are summarized as follows:
	\begin{itemize}
		\item We systematically re-evaluate existing SAR-GCD frameworks with DINOv2 and establish a DINOv2-based benchmark, which verifies the critical role of strong semantic prior in SAR-GCD.
		\item We propose the MDC, which explicitly characterizes the discrepancy between optical and SAR modalities in the frequency domain and provides a quantifiable frequency-domain descriptor for cross-modal alignment.
		\item We design the MCPT pre-training strategy, in which AFT and FER are integrated to perform spectral-guided prior transfer and band-wise feature refinement, thereby improving the adaptation of optical prior to the SAR domain.
		\item Extensive experiments on four mainstream datasets demonstrate the consistent performance of the proposed method and validate the effectiveness of spectral-guided discrepancy modeling for SAR-GCD.
	\end{itemize}
	
	\section{Related Work}
	\subsection{Generalized Category Discovery}
	Existing GCD methods can be broadly divided into two categories: non-parametric classifier methods and parametric classifier methods.
	
	\textbf{Non-parametric Classifier Methods.}
	These methods typically rely solely on a feature encoder and organize the feature space through contrastive learning, clustering, or graph-based modeling, upon which novel category discovery is performed. Early studies \cite{2_0_1,2_0_2} established the basic paradigm of combining contrastive representation learning with semi-supervised clustering \cite{2_0_3}. Subsequent research mainly focused on improving clustering quality from different perspectives, including latent hierarchical semantic mining \cite{2_0_4}, reliable cross-instance positive relation construction \cite{2_0_5}, intra-class distribution modeling \cite{2_0_6,2_0_7}, discriminative feature aggregation \cite{2_0_8,2_0_9,2_0_10}, and balanced clustering constraints \cite{2_0_11}. The advantage of this line lies in its openness, since it does not depend on an explicit classifier head. However, because the encoder learning process is inherently decoupled from the final category partition, its performance is highly sensitive to the stability of clustering guidance. Once the clustering structure on unlabeled data becomes unreliable, latent intra-class samples can be erroneously separated in the feature space.
	
	\textbf{Parametric Classifier Methods.}
	Parametric approaches introduce an explicit classifier head and jointly optimize the encoder and classifier using ground-truth labels for known classes together with pseudo-labels for novel classes, which usually yields stronger discriminative performance and has therefore become the mainstream paradigm. Representative studies in this line mainly address two issues: bias toward known classes and noisy pseudo-labels. Existing analyses \cite{2_0_1,2_0_2,2_1_1} have shown that the overfitting tendency of parametric GCD models is closely related to pseudo-label quality. Based on this observation, SimGCD \cite{2_1_1} employs self-distillation to alleviate prediction bias toward known classes. Follow-up works further improve pseudo-label reliability from the perspectives of active sample selection \cite{2_1_5} and adaptive confidence-based thresholding \cite{2_1_6}. Nevertheless, the effectiveness of these methods still depends heavily on pseudo-label quality, and once noise accumulates, confirmation bias can continue to propagate during training.
	
	Overall, whether improving clustering mechanisms in non-parametric methods or enhancing pseudo-label quality in parametric methods, existing GCD approaches fundamentally rely on strong feature representations. In the optical domain, such representations are often supported by semantic prior from large-scale pretrained vision models. In the SAR domain, however, this prerequisite becomes much more fragile due to complex scattering characteristics, substantial modality discrepancy, and limited annotations. Therefore, for SAR-GCD, the central challenge is no longer only the design of better clustering or pseudo-labeling strategies, but also the reliable transfer of strong pretrained optical prior into the heterogeneous SAR feature space.
	
	\subsection{Domain Adaptation}
	Domain adaptation methods aim to reduce the modality gap between optical and SAR imagery and thus facilitate cross-modal knowledge transfer. Existing studies can be roughly grouped into three directions. The first direction operates at the pixel level by constructing intermediate or transitional domains, for example through speckle-noise injection or texture-statistics simulation, so as to alleviate the abrupt appearance shift from optical images to SAR data \cite{2_2_1, 2_2_4, 2_2_5}. The second direction performs alignment in the feature space, where adversarial learning, global--local constraints, or instance-level consistency objectives are introduced to bring heterogeneous modalities closer in the semantic space \cite{2_2_2, 2_2_6}. The third direction adopts progressive transfer strategies, decomposing the transfer process into multiple stages, such as from natural images to optical remote sensing imagery and then to SAR, in order to reduce the difficulty of cross-modal adaptation \cite{2_2_3, 2_2_7}.
	
	Although these methods have shown effectiveness in facilitating cross-modal transfer, they generally treat modality discrepancy as an optimization target to be minimized and rely primarily on implicit end-to-end alignment. In other words, most existing methods focus on how to reduce the discrepancy, but do not explicitly characterize what structural form this discrepancy takes or how it relates to imaging characteristics. As a result, optical prior can only be transferred to the SAR domain in a coarse and weakly constrained manner, making it difficult to achieve fine-grained and interpretable adaptation. For the SAR-GCD task considered in this work, such implicitly aligned transfer is still insufficient to stably support both known-class recognition and novel-class discovery. This limitation motivates our method to explicitly model optical-SAR discrepancy from a frequency-domain perspective and to use it for interpretable cross-modal prior transfer.

	\section{Methodology}
	\subsection{Modality Discrepancy Curve: Frequency-Domain Discrepancy Representation}
	\subsubsection{Motivation and Design Rationale}
	SAR-GCD relies on semantic prior learned from optical data. Direct transfer of the prior is limited by the modality gap between optical and SAR imagery. Existing domain adaptation methods typically enforce alignment in the feature space and treat this gap as an implicit optimization target. Such strategies do not explicitly characterize the structure of the discrepancy in relation to imaging characteristics such as radiometric distortion, geometric variation, and speckle noise.
	
	To address this limitation, we introduce the MDC as the core component that governs the proposed framework. The MDC models the cross-modal discrepancy in the frequency domain, where modality differences exhibit structured energy distributions. Instead of relying on implicit alignment, the MDC provides a quantitative descriptor that captures the deviation between SAR and the optical reference across frequency bands.
	
	Concretely, the MDC is derived from the residual spectrum between SAR and optical imagery. As optical images provide more stable structural information and stronger generic visual prior, they serve as the reference anchor for discrepancy measurement. By computing frequency-domain residuals with respect to this reference, shared components are suppressed while modality-specific variations are emphasized. This process transforms spatial differences into structured spectral energy variations. As a result, the modality gap is represented as a frequency-dependent descriptor that facilitates subsequent learning.
	
	\subsubsection{MDC Formulation}
	The MDC serves as the core discrepancy representation in our method, modeling the cross-modal difference between optical and SAR imagery in the frequency domain. As shown in Fig. \ref{fig_3_1}, the formulation takes a spatially aligned optical image $x_{\text{EO}} \in \mathbb{R}^{H \times W \times 3}$ and a SAR image $x_{\text{SAR}} \in \mathbb{R}^{H \times W \times 1}$ as input, and outputs the discrepancy curve $C(f) \in \mathbb{R}^B$.
	
	The formulation builds upon residual computation by taking the optical image as the reference. Specifically, the residual image is defined as $r = x_{\text{SAR}} - x_{\text{EO}}$, where $x_{\text{EO}}$ is converted into a single channel. This operation removes shared components across modalities and isolates the deviation of SAR from the optical reference. Then, we transform the residual into the frequency domain to further characterize this discrepancy. The Fourier transform $\mathcal{F}$ is applied to obtain the spectral energy:
	\begin{align}
		E_r = |\mathcal{F}(r)|^2, \quad E_{x_{\text{EO}}} = |\mathcal{F}(x_{\text{EO}})|^2,
	\end{align}
	to characterize this discrepancy across frequencies, the spectrum is partitioned using Gaussian masks $\{M_{\text{MDC}}^i\}_{i=1}^B$ comprising $B$ bands. The masks follow a non-uniform design. Low-frequency regions are densely sampled to preserve structural information, whereas high-frequency regions are sparsely sampled to suppress speckle noise. Under this design, the central frequency $\mu_{\text{MDC}}^i$ and bandwidth $\sigma_{\text{MDC}}^i$ increase with the band index. Each mask is defined as:
	\begin{align}
		M^{i}_{\text{MDC}} = \exp\left( - \frac{(d - \mu^{i}_{\text{MDC}})^2}{2(\sigma^{i}_{\text{MDC}})^2} \right),
	\end{align}
	where $d$ is the distance to the DC component in the frequency domain. Utilizing the aforementioned pre-defined Gaussian masks $M_{\text{MDC}}^i$, the band-wise spectral energy is computed as:
	\begin{align}
		E_r^i = M_{\text{MDC}}^i \odot E_r, \quad E_{x_{\text{EO}}}^i = M_{\text{MDC}}^i \odot E_{x_{\text{EO}}},
	\end{align}
	where $\odot$ denotes element-wise multiplication. Building upon this foundation, to achieve explicit modality discrepancy quantification relative to the optical reference, the modality discrepancy intensity $R_i$ for the $i$-th sampled frequency band is formulated as the ratio of the residual energy to the optical reference energy in that specific band:
	\begin{figure}[h]
		\centering
		\includegraphics[width=1\linewidth]{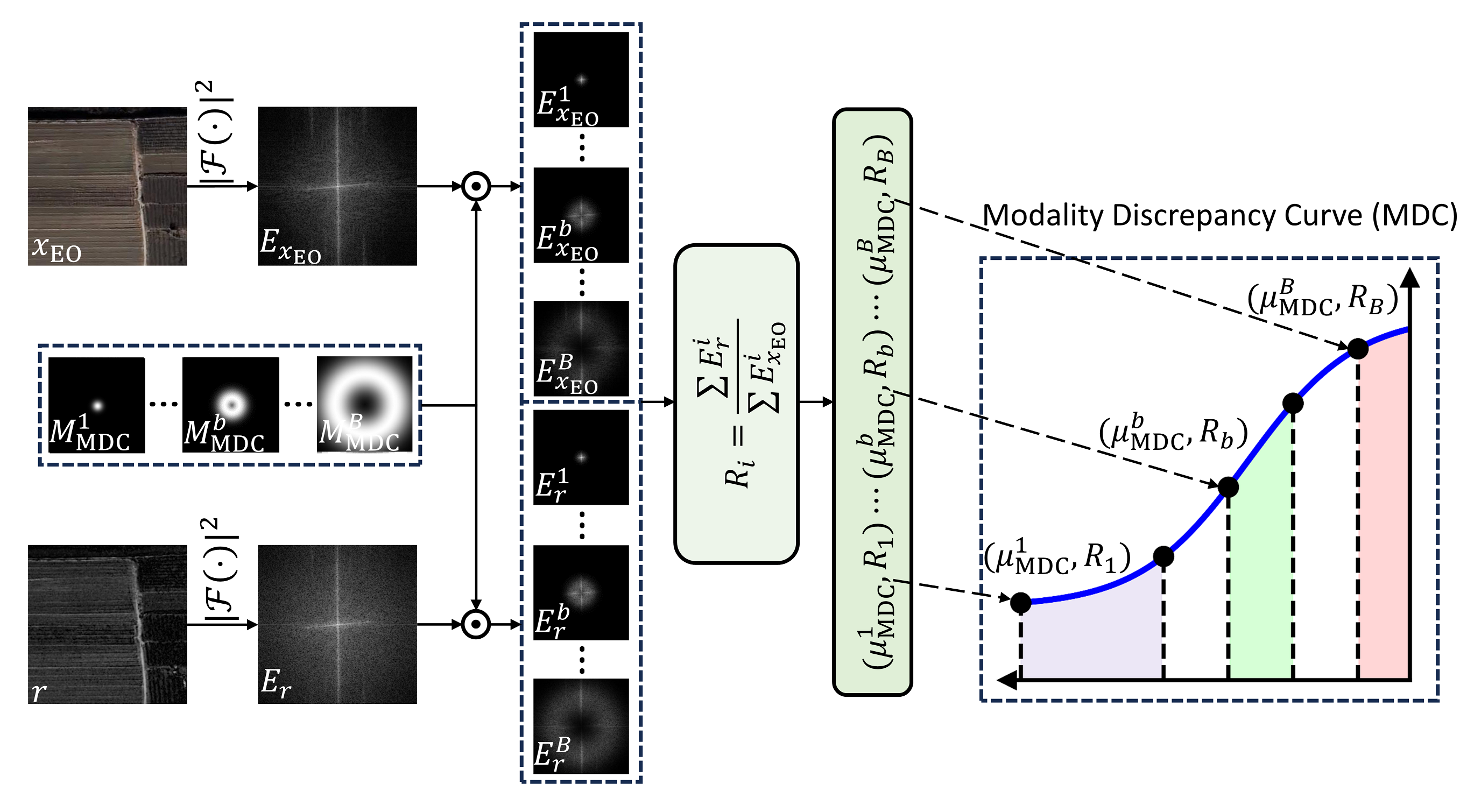}
		\caption{The Modal Discrepancy Curve (MDC) formulation process.}
		\label{fig_3_1}
	\end{figure}
	\begin{align}
		R_i = \frac{\sum E_r^i}{\sum E_{x_{\text{EO}}}^i},
	\end{align}
	normalizing residual energy by the optical reference energy reduces the influence of scene variations and makes discrepancy intensities comparable across frequency bands and samples. Mapping each frequency band energy ratio to the corresponding Gaussian mask central position yields the final vector representation $C(f)$ of the MDC, with $f$ representing the frequency from low to high. In summary, by establishing a residual spectrum quantification paradigm anchored on the optical benchmark, the MDC successfully converts the complex modality gap into a modality discrepancy descriptor in the frequency domain.

	\subsection{MCPT: MDC-guided Cross-modal Prior Transfer}
	\subsubsection{Overall Framework}
	The proposed framework is built upon the MDC, which explicitly characterizes the modal discrepancy and provides the basis for MCPT to guide the transfer of optical prior to the SAR domain. Instead of treating the modality gap as an implicit alignment target, the framework uses the MDC as an explicit frequency-domain representation that drives the entire transfer process.
	
	Based on this quantitative mechanism, MCPT adopts a dual-branch scheme in which both branches undergo feature refinement. Despite this structural symmetry, the alignment is inherently asymmetric, where optical features serve as semantic anchors that provide consistent guidance, while SAR features are adaptively aligned to them. In this process, the MDC specifies the deviation of the SAR modality from the optical reference across frequency bands, allowing the model to focus on structural shifts and noise patterns in different spectral regions and enabling targeted feature compensation.
	
\begin{figure*}[!h]
		\centering
		\includegraphics[width=0.9 \textwidth]{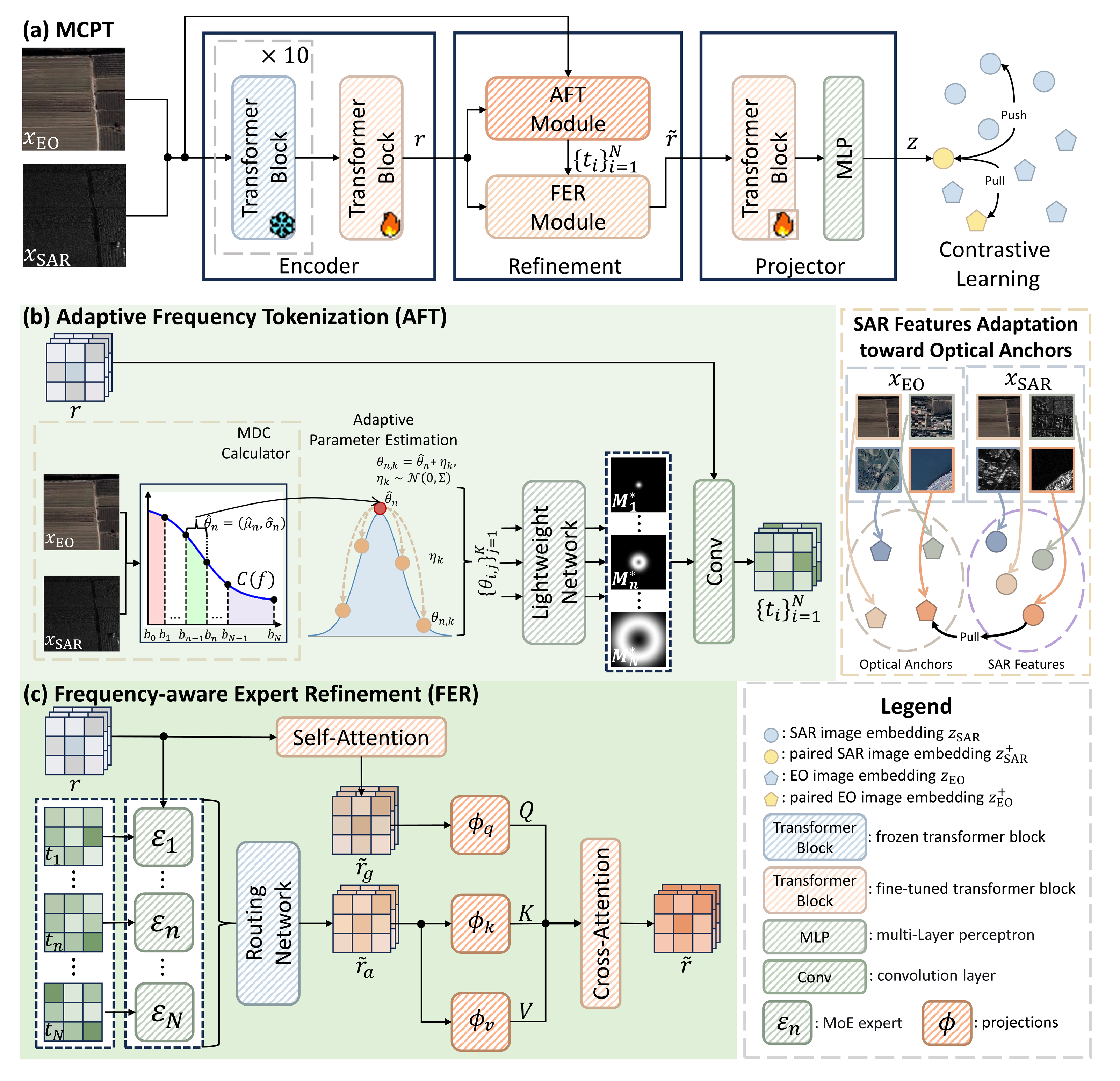}
		\caption{(a) The overall framework of the proposed MCPT. (b) Construction of the MDC and its discretization into spectral tokens via the AFT module. (c) Token-wise FcE-based feature refinement within the FER module.}
		\label{fig_3_2} 
	\end{figure*}

	The overall framework, as illustrated in Fig. \ref{fig_3_2}(a), employs the DINOv2 architecture equipped with 12 Transformer blocks as the shared encoder, taking paired optical-SAR images $\{x_{\text{EO}}, x_{\text{SAR}}\}$ as input. The paired data ${x_{\text{EO}}, x_{\text{SAR}}}$ are initially fed into the first 11 layers of the network $\{f_{\theta_\text{e}}^i\}_{i=1}^{11}$, producing intermediate representations $\{r_{\text{EO}}, r_{\text{SAR}}\}$ that encapsulate high-level semantic information. Thereafter, the learnable auxiliary modules AFT and FER are incorporated between the 11-th and 12-th layers to execute feature refinement. Following the processing by these auxiliary components, the resulting refined features $\{\widetilde{r}_{\text{EO}}, \widetilde{r}_{\text{SAR}}\}$ are fed into the final encoder layer $f_{\theta_\text{e}}^{12}$, mapping them to embeddings $\{z_{\text{EO}}, z_{\text{SAR}}\}$. In the final stage, a contrastive learning mechanism is employed to drive the SAR representations to adaptively align with the stable optical distribution, accomplishing the transfer of prior.
	
	Through this mechanism, the MDC is first converted into a network-compatible representation via AFT, producing learnable guidance tokens. These tokens are subsequently utilized by FER for feature refinement, forming a coherent MDC-driven process. Although the proposed framework comprises multiple components, it follows a unified MDC-driven design that translates explicit frequency-domain discrepancy into feature-space adaptation. Accordingly, AFT and FER are not independent architectural contributions but successive functional realizations that enable the MDC to influence feature learning.
	\subsubsection{Adaptive Frequency Tokenization}
	The MDC provides a structured description of cross-modal discrepancy in the frequency domain. To make this description usable in the network, it is converted into learnable guidance signals through the AFT. Therefore, this module defines the translation of the modeled discrepancy into representations that guide feature refinement. As such, AFT serves as a necessary modality bridge between spectral discrepancy representation and spatial feature learning.
	
	As illustrated in Fig. \ref{fig_3_2}(b), the curve $C(f)$ is partitioned into $N$ contiguous regions according to equal energy, where each region corresponds to a frequency band. For the $i$-th region, the initial Gaussian parameters $\hat{\theta}_i = (\hat{\mu}_i, \hat{\sigma}_i)$ are defined such that $\hat{\mu}_i$ represents the center of this region and the value of $\hat{\sigma}_i$ is proportional to its width. During this process, despite providing a stable initialization under the MDC distribution, equal-energy partition is still constrained by deterministic frequency parameter learning, which tends to collapse toward dominant bands. To overcome this limitation, an adaptive parameter estimation mechanism with stochastic parameter exploration is introduced in the parameter space \cite{3_0_1}. A local Gaussian distribution is constructed based on $\hat{\theta}_i$, from which $K$ candidate parameters $\{\theta_{i,j}\}_{j=1}^K = \{(\mu_{i,j}, \sigma_{i,j})\}_{j=1}^K$ are sampled via local Gaussian perturbation-based sampling. This process refines the band boundaries around informative frequency regions. A lightweight mapping $\varphi(\cdot)$ assigns weights to these candidates:
	\begin{align}
		\{w_{i,j}\}_{j=1}^K = \mathrm{softmax}(\varphi(\{\theta_{i,j}\}_{j=1}^K)),
	\end{align}
	which are used to obtain refined parameters $\theta_i^\ast = (\mu_i^\ast, \sigma_i^\ast)$:
	\begin{align}
		\mu_i^\ast = \sum_{j=1}^{K} w_{i,j} \cdot \mu_{i,j}, \quad \sigma_i^\ast = \sum_{j=1}^{K} w_{i,j} \cdot \sigma_{i,j},
	\end{align}
	thereafter, the parameters $\theta_i^\ast$ are leveraged to construct the adaptive ring-shaped Gaussian mask $M_i^\ast$ for the $i$-th frequency band:
	\begin{align}
		M_i^\ast = \exp\left(-\frac{(d - \mu_i^\ast)^2}{2(\sigma_i^\ast)^2}\right),
	\end{align}
	the AFT combines the features and the mask set $\{M_i^\ast\}_{i=1}^N$ to yield a group of spectral tokens $\{t_i\}_{i=1}^N$. The token $t_i$ for the $i$-th frequency band is defined as:
	\begin{align}
		t_i = \text{conv}_i(\mathcal{F}^{-1}(\mathcal{F}(\text{conv}_{\text{s}}(r)) \odot M_i^\ast)),
	\end{align}
	where $\text{conv}_{\text{s}}$ denotes a shared convolution, $\text{conv}_i$ denotes a specific convolution tailored for the $i$-th frequency band, and $\mathcal{F}^{-1}$ denotes the inverse Fourier transform. 
	
	Through this process, the MDC is transformed from a static curve into a set of tokens that encode discrepancy patterns across frequency bands. These tokens preserve the band-wise discrepancy information encoded in the MDC and provide guidance signals that can be directly used for subsequent feature refinement.
	
	\subsubsection{Frequency-aware Expert Refinement}
	FER leverages the spectral tokens generated by AFT to refine feature representations in a band-wise manner, where each token encodes discrepancy information derived from the MDC.
	
	In general, different frequency regions are associated with different modality discrepancy characteristics, which cannot be effectively modeled by a single shared transformation. To address this, a Frequency-conditioned Experts (FcE) mechanism is employed to process different frequency bands separately, enabling band-specific modulation through a lightweight expert design. As shown in Fig. \ref{fig_3_2}(c), given the input feature $r$ and the token set $\{t_i\}_{i=1}^N$, each token $t_i$ is routed to a corresponding expert $\varepsilon_i = (\phi_{\text{q}_i}, \phi_{\text{k}_i}, \phi_{\text{v}_i})$ to perform band-specific refinement. Concretely, based on the assigned guidance signal $t_i$, each expert performs band-specific compensation on the input features $r$ via a cross-attention mechanism to obtain the initial refined feature $\widetilde{r}_{\text{o},i}$:
	\begin{align}
		\widetilde{r}_{\text{o},i} = \mathrm{softmax}\left(\frac{\phi_{\text{q}_i}(r)(\phi_{\text{k}_i}(t_i))^{\mathrm{T}}}{\sqrt{d_\text{k}}}\right)\phi_{\text{v}_i}(t_i),
	\end{align}
	where $\phi_{\text{q}_\ast}$, $\phi_{\text{k}_\ast}$, and $\phi_{\text{v}_\ast}$ denote linear projections and $d_\text{k}$ is the dimension of the Key vector.
	
	Following the completion of the aforementioned local processing, an aggregation of the independent output results from all experts is required. In this step, a routing network $G(\cdot)$ evaluates the entire token set $\{t_i\}_{i=1}^N$ to compute the fusion weight vector $\{w_i\}_{i=1}^N$:
	\begin{align}
		\{w_i\}_{i=1}^N = \mathrm{softmax}(G(\{t_i\}_{i=1}^N)),
	\end{align}
	then, a weighted summation is performed on the outputs of all experts according to this weight vector, producing the comprehensive refined feature $\widetilde{r}_{\text{a}}$ that integrates multi-band discrepancy compensation information:
	\begin{align}
		\widetilde{r}_{\text{a}} = \sum_{i=1}^{N} w_i \widetilde{r}_{\text{o},i},
	\end{align}
	since band-wise refinement may disrupt global semantic coherence, a global attention branch is introduced to preserve cross-band consistency. To this end, a global semantic representation is first obtained from $r$ via self-attention, and its attention weights are used to modulate the discrepancy-aware feature $\widetilde{r}_{\text{a}}$, thereby preserving global semantic coherence during band-wise refinement:
	\begin{align}
		\widetilde{r} = \mathrm{softmax}\left(\frac{\phi_\text{q}(\widetilde{r}_{\text{g}})(\phi_\text{k}(\widetilde{r}_{\text{g}}))^{\mathrm{T}}}{\sqrt{d_\text{k}}}\right)\phi_\text{v}(\widetilde{r}_{\text{a}}),
	\end{align}
	to summarize, the modal discrepancy encoded in the MDC is leveraged by FER to guide feature refinement across frequency bands while preserving global semantics. The resulting feature $\widetilde{r}$ integrates band-wise discrepancy-aware information derived from the MDC and serves as input for subsequent alignment.
	
	\subsubsection{Contrastive Alignment and Parameter Internalization}
	Based on the MDC-guided refinement, in which multi-band discrepancy-aware refinement is achieved through AFT and FER, contrastive learning is further applied to the refined embeddings, enabling the shared encoder to internalize cross-modal alignment patterns into its parameters.
	
	To achieve this, a joint objective combining cross-modal symmetric loss and unsupervised contrastive loss is defined. The unidirectional symmetric loss $L_{\text{SAR} \rightarrow \text{EO}, i}$ from SAR to optical is computed as:
	\begin{align}
		L_{\text{SAR} \to \text{EO}, i} &= - \left( \frac{1}{2} \log \frac{\exp(z_{\text{SAR}}^i \cdot z_{\text{EO}}^i / \tau)}{\sum_{j \in J_1} \exp(z_{\text{SAR}}^i \cdot z_{\text{EO}}^j / \tau)} \right. \nonumber \\
		&\quad \left. + \frac{1}{2} \log \frac{\exp(z_{\text{SAR}}^i \cdot z_{\text{EO}}^{i,+} / \tau)}{\sum_{j \in J_1} \exp(z_{\text{SAR}}^i \cdot z_{\text{EO}}^j / \tau)} \right),
	\end{align}
	where $z_{\text{SAR}}^i$ and $z_{\text{EO}}^i$ denote the $i$-th SAR and optical embeddings within the current batch, $z_{\text{EO}}^{i,+}$ denotes the augmented embedding of the $i$-th optical image, $J_1$ denotes the index set of all optical embeddings in the batch (including both original and augmented samples, totaling $2N$), and $\tau$ denotes the temperature parameter. The reverse loss $L_{\text{EO}\to\text{SAR}}$ is defined similarly, and the symmetric loss is:
	\begin{align}
		L_{\text{sym}}=\frac{1}{2N_{\text{B}}}\sum_{i=1}^{N_{\text{B}}}\left(L_{\text{SAR}\rightarrow\text{EO},i}+L_{\text{EO}\rightarrow\text{SAR},i}\right),
	\end{align}
	where $N_{\text{B}}$ denotes the batch size. By minimizing the distance between heterogeneous modalities in the feature space, this loss promotes the alignment of SAR features with stable optical anchors. Furthermore, to preserve the inherent semantic discriminability of the features in the cross-modal alignment, an unsupervised contrastive loss $L_{\text{unsup}}$ is introduced:
	\begin{align}
		L_{\text{unsup}}=-\frac{1}{N_{\text{B}}}\sum_{i=1}^{N_{\text{B}}}\log\frac{\exp\left(z^i\cdot z^{i,+}/\tau\right)}{\sum_{j\in J_2,j\neq i}\exp\left(z^i\cdot z^j/\tau\right)},
	\end{align}
	where $z^i$ denotes the embedding of a given sample view, $z^{i,+}$ denotes its augmented positive counterpart, $J_2$ denotes the index set of all embeddings in the batch. A weighted summation of the cross-modal symmetric loss and the unsupervised contrastive loss yields the total loss $L_{\text{total}}$:
	\begin{align}
		L_{\text{total}}=(1-\lambda)L_{\text{sym}}+\lambda L_{\text{unsup}},
	\end{align}
	where $\lambda$ defines the weighting coefficient.
	
	In this context, the MDC-guided learning process establishes a dual mechanism for feature enhancement and alignment, which integrates explicit spectral guidance with implicit parameter internalization. Under explicit guidance, the MDC is translated into spectral tokens through AFT, and these tokens are applied via FER to perform band-wise feature refinement, enabling the network to capture structured cross-modal discrepancy patterns. Concurrently, from the implicit internalization perspective, optimization constraints are imposed in the embedding space through the joint contrastive loss, encouraging SAR representations to match stable optical prior anchors.
	
	More importantly, through continuous backpropagation, a dynamic interaction is established between these two components. The discrepancy-aware features obtained under MDC guidance provide discriminative representations for contrastive optimization, while the gradients from contrastive learning further enhance the representation capacity of the encoder, resulting in features that better align with the MDC-defined discrepancy patterns. This closed-loop learning process of mutual reinforcement accelerates the adaptive adjustment of SAR features, allowing the backbone network to efficiently internalize the optical-to-SAR cross-modal alignment pattern into its parameters.
	
	Ultimately, as the alignment is progressively internalized, the backbone network learns to transfer optical prior independently, with AFT and FER serving only as auxiliary components during pre-training. Finally, these auxiliary modules can be removed during fine-tuning and inference, thereby maintaining model efficiency and eliminating reliance on paired data.
	
	\subsection{Two-Stage Optimization Strategy}
	To deploy the MDC-driven feature representation for the SAR-GCD task, a two-stage optimization paradigm is adopted. This design serves as the transition from MDC-guided feature construction to downstream semantic-level target discovery. In this process, the first stage focuses on building effective representations under the MDC guidance, while the second stage applies these representations to the SAR-GCD task.
	
	During the MCPT pre-training phase, the objective is to transfer optical prior to the SAR domain under MDC guidance. Although the paired pre-training data are provided at the scene level rather than the target level, MCPT is designed to facilitate domain transfer by learning category-agnostic cross-modal discrepancy patterns, thereby enhancing SAR feature representations for downstream SAR-GCD tasks. The MCPT framework is employed to enable the encoder to learn SAR-adaptive representations by transferring optical prior under the guidance of the MDC, resulting in discriminative features for SAR images. Concretely, DINOv2 with a 12-layer Transformer structure is used as the shared encoder, and a partial parameter freezing strategy is applied. The first 10 layers of the encoder are frozen, while the 11th and 12th layers together with the AFT and FER modules are updated through backpropagation. Within the optimization process, the MDC is translated into guidance signals and applied to feature learning, while the joint loss function $L_{\text{total}}$ enforces alignment in the embedding space. After pre-training, the AFT and FER modules are removed, and the downstream model retains the original DINO-based GCD backbone, with the shared encoder enhanced through MDC-guided learning, thereby preserving its inference efficiency.
	
	In the fine-tuning stage for the SAR-GCD task, to translate the high-quality MDC-empowered representations acquired during pre-training into category discovery capabilities, the pre-trained DINOv2 encoder is incorporated into the ProtoGCD algorithm \cite{2_1_6} as the feature extractor and subsequently fine-tuned on an unpaired single-modal SAR-ATR dataset. The optimization objective is defined as $L_{\text{fine}} = L_{\text{cls}} + L_{\text{con}} + L_{\text{reg}}$, where $L_{\text{cls}}$ incorporates pseudo-label supervision, $L_{\text{con}}$ promotes intra-class compactness, and $L_{\text{reg}}$ includes entropy regularization and prototype separation.
	
	Through this two-stage process, the MDC-guided representation learning is first established and then deployed for high-dimensional category discovery, enabling superior SAR target recognition in the SAR-GCD setting.
	
	\begin{table}[t]
		\centering
		\caption{Statistics of the Four Datasets}
		\begin{tabular}{llllll}
			\toprule
			\multirow{2}{*}{Datasets} & \multicolumn{2}{c}{Classes} & \multicolumn{3}{c}{Sample Size} \\
			\cmidrule(lr){2-3} \cmidrule(lr){4-6}
			& Old   & New   & Labeled & Unlabeled & Test \\
			\midrule
			MSTAR & 6     & 4     & 808     & 1939      & 2425 \\
			SAMPLE & 7     & 3     & 285     & 651       & 409 \\
			FUSAR & 4     & 3     & 297     & 418       & 311 \\
			OpenSARShip & 3     & 2     & 220     & 385       & 262 \\
			\bottomrule
		\end{tabular}%
		\label{table_4_0_1}%
	\end{table}%

	\section{Experiments}
	\subsection{Experimental settings}
	\subsubsection{Datasets}
	Since the MCPT phase requires paired optical-SAR data to drive cross-modal knowledge transfer, pre-training is performed on the high-quality paired YESeg-OPT-SAR dataset \cite{4_0_1}. This dataset covers diverse scenes, such as buildings, roads, farmlands, vegetation, and water, and comprises 2,231 strictly registered SAR-optical image patch pairs. All images are uniformly cropped to a size of $256 \times 256$ pixels, corresponding to an actual ground coverage of $128 \times 128$ m with a high spatial resolution of 0.5 m.
	
	For the SAR target recognition task, experiments are conducted on the general datasets MSTAR \cite{4_0_2} and SAMPLE \cite{4_0_3}, alongside the more challenging long-tailed datasets FUSAR \cite{4_0_4} and OpenSARShip \cite{4_0_5}. Following standard experimental settings in the literature on GCD \cite{2_0_1}, a subset of categories in each dataset is designated as old classes, leaving the remainder as new classes. Subsequently, 50\% of the instances from the original old-class training set are randomly sampled to construct the labeled dataset $D_l$, with all remaining data constituting the unlabeled dataset $D_u$. To comprehensively assess model performance, we adopt two evaluation paradigms. Under the standard transductive setting, the model undergoes joint training on $D_l \cup D_u$ prior to evaluation on the unlabeled dataset $D_u$. Conversely, the inductive setting evaluates the fully trained model on an independent test dataset $D_{\text{test}}$. Detailed statistics for each dataset are summarized in Table \ref{table_4_0_1}.
	
	\subsubsection{Implementation Details}
	This work adopts a two-stage training paradigm comprising pre-training and fine-tuning, employing the ViT-B/16 architecture of DINOv2 \cite{4_0_9} as the backbone. During the pre-training phase, the number of predefined masks for the construction of MDC is set to 25. Concurrently, the number of frequency bands within the AFT module is uniformly established as 4, aligning with the number of experts in the FER module. Regarding the parameter optimization strategy, we selectively update the auxiliary modules (AFT and FER) alongside the final two layers of the backbone network, while freezing the remaining layers to preserve generalizable visual knowledge. Furthermore, the model is trained for 150 epochs with a batch size of 64 using the AdamW optimizer ($\beta_1 = 0.9, \beta_2 = 0.999$) at an initial learning rate of $1 \times 10^{-4}$ with cosine decay, and the loss weight coefficient $\lambda$ is set to 0.5. Upon the completion of pre-training, the auxiliary modules are discarded. Only the DINOv2 backbone, having encoded cross-modal knowledge, is retained for the SAR-GCD fine-tuning phase. In the fine-tuning stage, the pre-trained DINOv2 is incorporated into the SOTA GCD framework, ProtoGCD \cite{2_1_6}, to validate the effectiveness of the proposed approach. For a fair comparison with methods utilizing DINOv1 \cite{4_0_8}, comparative experiments are simultaneously conducted on the DINOv1 architecture. Moreover, consistent with previous experimental settings \cite{2_0_1}, the fine-tuning process solely updates the final layer of the backbone. The fine-tuning procedure is executed across 200 epochs with the batch size maintained at 128 and the initial learning rate set to 0.01, dynamically regulated via a cosine annealing strategy. All experiments are conducted on a single NVIDIA GeForce RTX 4090 GPU.
	
	\begin{table*}[htbp]
		\centering
		\caption{Overall Comparison Results on the Four Datasets}
		\begin{tabular}{llllllllllllll}
			\toprule
			\multirow{2}{*}{Backbone} & \multirow{2}{*}{Methods} & \multicolumn{3}{c}{MSTAR} & \multicolumn{3}{c}{SAMPLE} & \multicolumn{3}{c}{FUSAR} & \multicolumn{3}{c}{OpenSARShip} \\
			\cmidrule(lr){3-5} \cmidrule(lr){6-8} \cmidrule(lr){9-11} \cmidrule(lr){12-14}
			&       & All   & Old   & New   & All   & Old   & New   & All   & Old   & New   & All   & Old   & New \\
			\midrule
			\multirow{7}{*}{DINOv1} 
			& GCD       & 77.26 & 89.74 & 68.32 & 71.43 & 95.10 & 52.88 & 42.82 & 43.10 & 42.15 & 40.52 & 37.56 & \textbf{44.51} \\
			& SimGCD    & 77.77 & 76.89 & 78.41 & 80.80 & 95.80 & 69.04 & 50.72 & 53.54 & \textbf{43.80} & 40.78 & 52.04 & 25.61 \\
			& InfoSieve & 77.41 & 71.69 & 81.50 & 73.73 & 85.66 & 64.38 & 44.98 & 53.87 & 23.14 & 40.26 & 43.44 & 35.98 \\
			& CMS       & 81.85 & 82.82 & 81.15 & 81.72 & 97.55 & 69.32 & 49.28 & \textbf{61.95} & 18.18 & 42.60 & 46.15 & 37.80 \\
			& BKD-CL    & 80.06 & \textbf{93.45} & 71.42 & 81.26 & 98.25 & 67.95 & 48.33 & 56.23 & 28.93 & 42.86 & 46.15 & 38.41 \\
			& ProtoGCD  & 82.52 & 79.73 & 84.51 & 81.41 & \textbf{99.65} & 67.12 & 50.00 & 61.62 & 21.49 & 43.90 & 47.06 & 39.63 \\
			\cmidrule(lr){2-14}
			& Ours      & \textbf{90.77} & 90.98 & \textbf{90.62} & \textbf{86.94} & 98.95 & \textbf{77.53} & \textbf{51.91} & 56.23 & 41.32 & \textbf{45.19} & \textbf{52.94} & 34.76 \\
			\midrule
			\multirow{9}{*}{DINOv2} 
			& GCD       & 85.10 & 90.98 & 80.88 & 82.49 & 91.96 & 75.07 & 45.45 & 49.83 & 34.71 & 41.82 & 43.89 & 39.02 \\
			& SimGCD    & 87.73 & 90.48 & 85.75 & 82.03 & 94.06 & 72.60 & 50.24 & 53.87 & 41.32 & 41.82 & 50.23 & 30.49 \\
			& InfoSieve & 88.81 & 95.30 & 84.16 & 84.18 & 84.27 & 84.11 & 51.44 & 59.93 & 30.58 & 41.56 & 43.44 & 39.02 \\
			& CMS       & 86.33 & 97.16 & 78.58 & 88.48 & 95.10 & 83.29 & 50.24 & \textbf{63.97} & 16.53 & 42.08 & 46.61 & 35.98 \\
			& BKD-CL    & 87.93 & 92.09 & 84.96 & 89.25 & 97.55 & 82.74 & 51.44 & 58.25 & 34.71 & 43.90 & 48.42 & 37.80 \\
			& ProtoGCD  & 90.30 & 90.61 & 90.09 & 92.17 & 98.25 & 87.40 & 52.87 & 58.92 & 38.02 & 44.68 & 51.13 & 35.98 \\
			& ProtoGCD (w/ CM-PT)  & 90.77 & 91.35 & 90.35 & 92.47 & 94.76 & 90.68 & 53.35 & 60.27 & 36.36 & 45.45 & 54.30 & 33.54 \\
			\cmidrule(lr){2-14}
			& Ours      & \textbf{98.19} & \textbf{99.03} & \textbf{97.61} & \textbf{98.77} & \textbf{99.83} & \textbf{97.96} & \textbf{58.13} & 61.62 & \textbf{49.59} & \textbf{48.05} & \textbf{54.75} & \textbf{39.02} \\
			\bottomrule
		\end{tabular}%
		\label{table_4_1_1}%
	\end{table*}%
	
	\begin{table*}[htbp]
		\centering
		\caption{Inductive Evaluation Results on the Four Datasets}
		\begin{tabular}{lllllllllllllll}
			\toprule
			\multirow{2}{*}{Backbone} & \multirow{2}{*}{Methods} & \multicolumn{3}{c}{MSTAR} & \multicolumn{3}{c}{SAMPLE} & \multicolumn{3}{c}{FUSAR} & \multicolumn{3}{c}{OpenSARShip} \\
			\cmidrule(lr){3-5} \cmidrule(lr){6-8} \cmidrule(lr){9-11} \cmidrule(lr){12-14}
			&       & All   & Old   & New   & All   & Old   & New   & All   & Old   & New   & All   & Old   & New \\
			\midrule
			\multirow{7}{*}{DINOv1} 
			& GCD       & 75.46 & 83.52 & 64.31 & 68.22 & 77.20 & 54.09 & 41.48 & 39.69 & \textbf{50.00} & 40.46 & 40.31 & 40.85 \\
			& SimGCD    & 79.71 & 90.13 & 65.30 & 84.60 & 94.80 & 68.55 & 50.80 & 53.70 & 37.04 & 41.22 & 47.64 & 23.94 \\
			& InfoSieve & 75.05 & 83.81 & 62.93 & 74.33 & 81.20 & 63.52 & 46.95 & 52.14 & 22.22 & 40.84 & 37.70 & \textbf{49.30} \\
			& CMS       & 81.15 & \textbf{97.44} & 58.60 & 83.62 & 93.60 & 67.92 & 49.20 & 55.64 & 18.52 & 41.60 & 51.31 & 15.49 \\
			& BKD-CL    & 84.04 & 95.67 & 67.94 & 86.06 & 97.20 & 68.55 & 50.16 & 54.09 & 31.48 & 43.13 & 48.69 & 28.17 \\
			& ProtoGCD  & 84.95 & 91.26 & 76.20 & 85.09 & \textbf{98.00} & 64.78 & 51.13 & \textbf{58.37} & 16.67 & 42.37 & 47.64 & 28.17 \\
			\cmidrule(lr){2-14}
			& Ours      & \textbf{93.36} & 94.32 & \textbf{92.04} & \textbf{89.73} & 97.20 & \textbf{77.99} & \textbf{52.41} & 54.47 & 42.59 & \textbf{43.51} & \textbf{51.31} & 22.54 \\
			\midrule
			\multirow{8}{*}{DINOv2} 
			& GCD       & 84.54 & 89.35 & 77.88 & 83.37 & 91.60 & 70.44 & 43.73 & 46.30 & 31.48 & 43.89 & 45.55 & \textbf{39.44} \\
			& SimGCD    & 89.44 & 91.76 & 86.23 & 84.11 & 92.40 & 71.07 & 52.41 & 55.64 & 37.04 & 43.51 & 48.69 & 29.58 \\
			& InfoSieve & 85.90 & 93.96 & 74.73 & 82.40 & 78.40 & 88.68 & 50.16 & 53.31 & 35.19 & 40.84 & 47.12 & 23.94 \\
			& CMS       & 85.57 & 96.38 & 70.60 & 83.13 & 91.20 & 70.44 & 51.77 & 55.64 & 33.33 & 44.27 & 49.74 & 29.58 \\
			& BKD-CL    & 87.30 & 92.61 & 79.94 & 89.00 & 94.40 & 80.50 & 51.13 & 56.03 & 27.78 & 43.51 & \textbf{54.97} & 12.68 \\
			& ProtoGCD  & 90.14 & 90.91 & 89.09 & 92.67 & 97.60 & 84.91 & 52.73 & 57.20 & 31.48 & 43.89 & 48.17 & 32.39 \\
			& ProtoGCD (w/ CM-PT) & 91.38 & 91.83 & 90.76 & 91.20 & 93.60 & 87.42 & 53.05 & \textbf{58.75} & 25.93 & 44.66 & 53.93 & 19.72 \\
			\cmidrule(lr){2-14}
			& Ours      & \textbf{97.98} & \textbf{98.56} & \textbf{97.16} & \textbf{97.80} & \textbf{98.43} & \textbf{96.80} & \textbf{55.63} & 57.98 & \textbf{44.44} & \textbf{46.56} & 50.79 & 35.21 \\
			\bottomrule
		\end{tabular}%
		\label{table_4_1_2}%
	\end{table*}%
	
	To comprehensively evaluate the effectiveness and superiority of the proposed method, this study selects several representative existing GCD approaches as comparative baselines, broadly encompassing the two mainstream paradigms based on non-parametric and parametric classifiers. Specifically, the non-parametric classifier baselines include GCD \cite{2_0_1}, CMS \cite{2_0_8}, and InfoSieve \cite{4_0_6}, whereas the parametric classifier baselines incorporate SimGCD \cite{2_1_1}, ProtoGCD \cite{2_1_6}, and the SAR-tailored BKD-CL \cite{4_0_7}. All comparative experiments strictly adhere to the standard experimental protocols within the GCD domain. To reduce variance introduced by random partitioning, we use a fixed random seed for dataset splitting and report a consistent evaluation protocol across all methods. In addition, a pre-training baseline, ProtoGCD (w/ CM-PT), is introduced to disentangle the effect of the MDC-guided domain transfer from that of the additional paired data.
	
	\subsection{GCD Performance Comparison}
	\subsubsection{Comparisons with State-of-the-Art}
	To comprehensively evaluate the effectiveness of the proposed approach, we compare our methodology with mainstream GCD baselines (GCD \cite{2_0_1}, CMS \cite{2_0_8}, InfoSieve \cite{4_0_6}, SimGCD \cite{2_1_1}, BKD-CL \cite{4_0_7}, and ProtoGCD \cite{2_1_6}) under the transductive setting. The corresponding results are summarized in Table \ref{table_4_1_1}, where the optimal values are highlighted in bold. Empirical evaluations show that our method achieves consistent performance improvements across diverse backbone architectures and datasets. Specifically, when evaluated on the general SAR target recognition dataset MSTAR using the DINOv2 backbone, our method yields a $7.89\%$ improvement in average accuracy over the recent SOTA approach, ProtoGCD. Concurrently, a substantial performance margin of $5.26\%$ over the best existing baseline is observed on the long-tailed FUSAR dataset. More crucially, focusing on new class recognition, the presented methodology attains the highest accuracy across nearly all datasets with the DINOv2 backbone. These results indicate that MCPT improves the adaptation of optical prior to the SAR domain and is particularly beneficial for novel-class recognition. It is also observed that merely substituting the backbone architecture from DINOv1 to DINOv2 delivers exceptional accuracy enhancements across all evaluated methods, demonstrating the importance of strong pretrained visual representations.
	
	Moreover, to disentangle the effect of paired optical–SAR data leveraged during pre-training from that of the proposed MDC-guided mechanism, we construct a dedicated baseline, ProtoGCD (w/ CM-PT), in which the model is pre-trained using paired data without MDC guidance. The consistent gains over ProtoGCD (w/ CM-PT) suggest that the improvement is not primarily due to the additional paired data, but rather to the proposed MCPT framework.
	
	\subsubsection{Inductive Evaluation}
	Conventional GCD paradigms typically adopt the transductive evaluation strategy, wherein performance validation is directly executed on the unlabeled datasets utilized during the training phase. Striving to extensively assess the generalization capability in real-world open scenarios, the comparative experiments are further extended to encompass an inductive evaluation strategy, which conducts validation on a test set strictly isolated from the training procedure. The results in Table \ref{table_4_1_2} demonstrate that the overall accuracy of all baseline methods under the inductive paradigm is universally inferior to their transductive counterparts due to the inherent invisibility of the test samples. Nevertheless, the proposed framework consistently maintains optimal performance. Concretely, with the DINOv2 backbone, our method yields a $7.84\%$ improvement in average accuracy on the MSTAR dataset relative to the recent SOTA approaches, as well as a $2.9\%$ performance improvement on the FUSAR dataset. These empirical observations indicate that transferring optical prior to the SAR domain enables the network to extract more discriminative features and improves generalization to unseen SAR targets.
		\begin{figure}[t]
		\centering
		\includegraphics[width=1\linewidth]{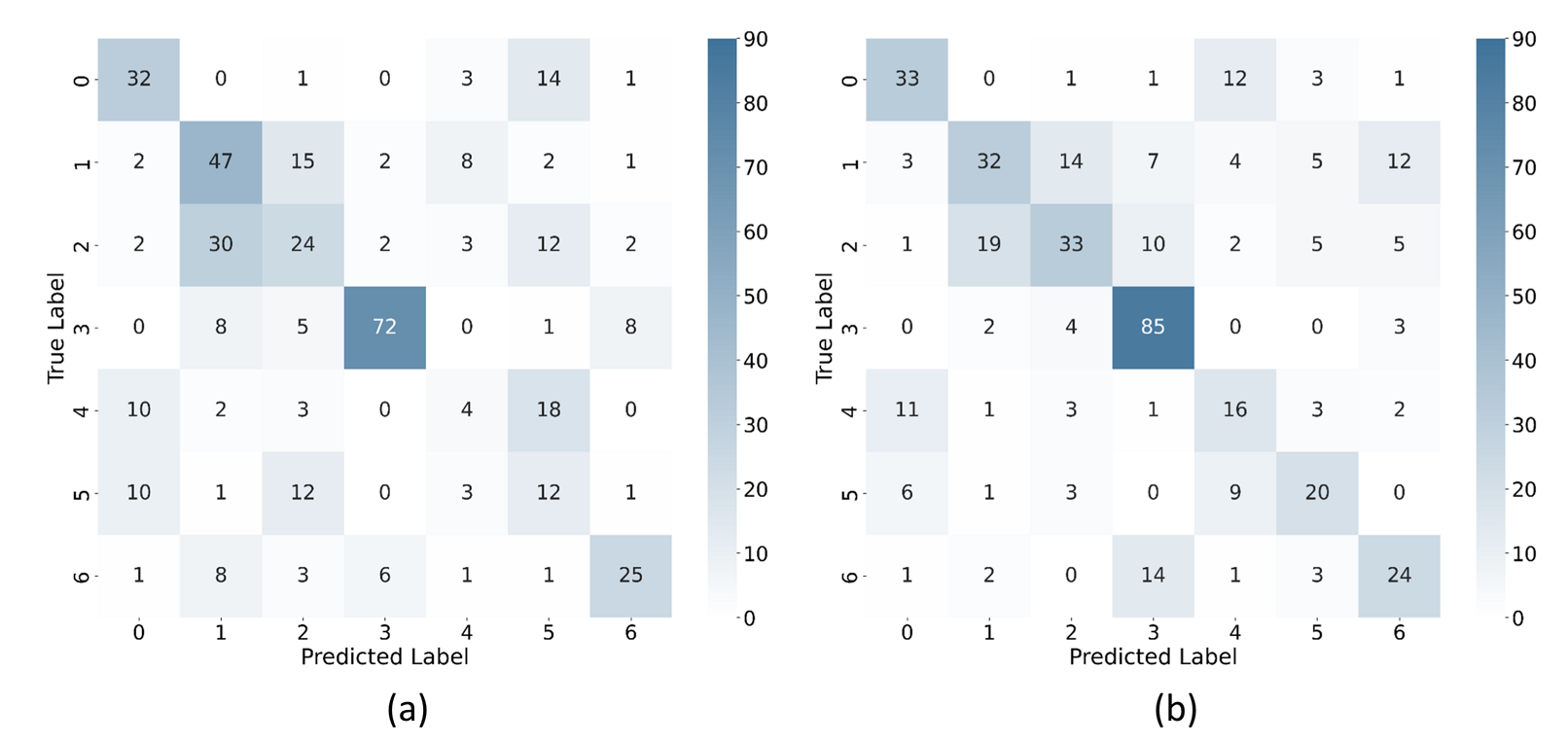}
		\caption{Confusion matrices on the long-tail FUSAR dataset with DINOv2. (a) ProtoGCD. (b) Ours.}
		\label{fig_4_2_1}
	\end{figure}
	
	\subsubsection{Confusion Matrix Analysis on the Long-tail FUSAR Dataset}
	Performance on the challenging long-tail FUSAR dataset is further analyzed through the confusion matrices shown in Fig. \ref{fig_4_2_1}, where the classes are defined as follows: (0) Tanker, (1) Fishing, (2) Cargo Ship, (3) Bulk Carrier, (4) Tug, (5) Dredger, and (6) Container Ship. The confusion matrices show that our method improves classification performance for several challenging classes compared to ProtoGCD. For example, the number of Cargo Ship samples (class 2) misclassified as Fishing (class 1) decreases from 30 under ProtoGCD to 19 with our method. New tail classes such as Tug (class 4) and Dredger (class 5) still exhibit confusion, but the number of misclassified samples is substantially reduced compared to ProtoGCD. Although the improvement is less pronounced than that observed on the well-balanced MSTAR dataset, the proposed method still delivers consistent gains over the baseline on long-tail datasets. This behavior is attributed to the scarcity of novel-class samples and the lack of supervision, which make structurally similar ship categories difficult to distinguish, even with cross-modal prior.
	
	\begin{table}[!t]
	\centering
	\caption{Main Ablation Studies for the Proposed Method on the MSTAR Dataset with DINOv1.}
	\begin{tabular}{llllllll}
		\toprule
		\multicolumn{5}{c}{Model Components} & \multicolumn{3}{c}{MSTAR} \\
		\cmidrule(lr){1-5} \cmidrule(lr){6-8}
		MCPT & FER & FcE & MDC & APE & All & Old & New \\
		\midrule
		$\times$   & $\times$   & $\times$   & $\times$   & $\times$   & 82.52 & 79.73 & 84.51 \\
		\checkmark & $\times$   & $\times$   & $\times$   & $\times$   & 84.32 & 94.31 & 77.17 \\
		\checkmark & \checkmark & $\times$   & $\times$   & $\times$   & 85.40 & 95.30 & 78.32 \\
		\checkmark & \checkmark & \checkmark & $\times$   & $\times$   & 86.54 & 91.10 & 83.27 \\
		\checkmark & \checkmark & \checkmark & \checkmark & $\times$   & 88.86 & 92.95 & 85.93 \\
		\checkmark & \checkmark & \checkmark & \checkmark & \checkmark & 90.77 & 90.98 & 90.62 \\
		\bottomrule
	\end{tabular}%
	\label{table_4_3_1}%
\end{table}%

\begin{table}[!t]
	\centering
	\caption{Main Ablation Studies for the Proposed Method on the MSTAR Dataset with DINOv2.}
	\begin{tabular}{llllllll}
		\toprule
		\multicolumn{5}{c}{Model Components} & \multicolumn{3}{c}{MSTAR} \\
		\cmidrule(lr){1-5} \cmidrule(lr){6-8}
		MCPT & FER & FcE & MDC & APE & All & Old & New \\
		\midrule
		$\times$   & $\times$   & $\times$   & $\times$   & $\times$   & 90.30 & 90.61 & 90.09 \\
		\checkmark & $\times$   & $\times$   & $\times$   & $\times$   & 90.77 & 91.35 & 90.35 \\
		\checkmark & \checkmark & $\times$   & $\times$   & $\times$   & 91.18 & 91.10 & 91.24 \\
		\checkmark & \checkmark & \checkmark & $\times$   & $\times$   & 92.57 & 93.08 & 92.21 \\
		\checkmark & \checkmark & \checkmark & \checkmark & $\times$   & 95.93 & 96.54 & 95.49 \\
		\checkmark & \checkmark & \checkmark & \checkmark & \checkmark & 98.19 & 99.03 & 97.61 \\
		\bottomrule
	\end{tabular}%
	\label{table_4_3_2}%
\end{table}%
	
	\subsection{Ablation Studies}	
	We conduct ablation studies on the MSTAR dataset to validate the effectiveness of individual components in the proposed method. The quantitative results are shown in Table \ref{table_4_3_1} and Table \ref{table_4_3_2}, and configurations (a)–(f) are introduced to clearly illustrate the experimental settings. Concretely, the configurations are defined as follows: (a) the baseline model is directly fine-tuned on the downstream SAR-GCD task; (b) cross-modal feature alignment is achieved using optical–SAR paired data and contrastive learning, where configuration (b) isolates the gain from paired cross-modal pretraining without MDC-guided refinement; (c) the FcE is removed, and feature refinement is performed using the FER module, which reduces to a standard self-attention mechanism; (d) frequency-domain tokens are generated using a uniform frequency band partition strategy instead of the MDC, and feature refinement is performed in combination with the FcE; (e) frequency-disentangled feature refinement is performed by combining the MDC with the FcE; (f) adaptive parameter estimation is integrated to enhance the MDC.
	
	\begin{figure*}[!t]
		\centering
		\includegraphics[width=1\linewidth]{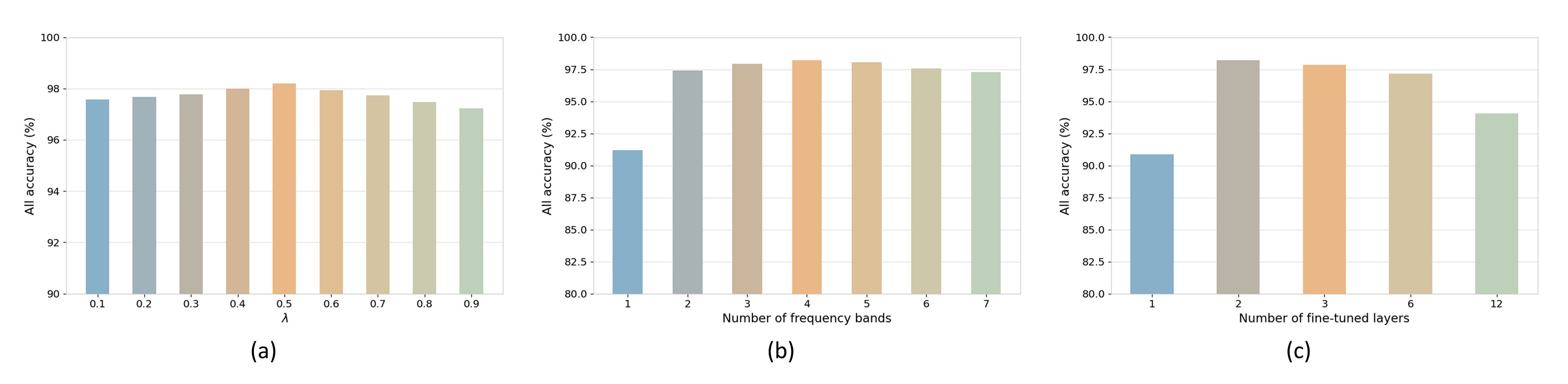}
		\caption{Parameter tuning studies of the proposed method on the MSTAR dataset with DINOv2. (a) Weight coefficient $\lambda$. (b) Frequency band counts $N$. (c) Fine-tuned layer counts $l$.}
		\label{fig_4_4_1}
	\end{figure*}
	
	As shown in Table \ref{table_4_3_1}, (b) achieves a performance improvement over (a), demonstrating the effectiveness of the domain transfer mechanism for SAR-GCD. The self-attention mechanism in (c) and the FcE in (d) merely strengthen the basic representation capability, leading to limited gains in recognition accuracy. Benefiting from the incorporation of the MDC in (e), the FER module significantly improves performance by performing feature refinement based on band-wise modality discrepancy information. In (f), adaptive parameter estimation transforms static curves into a learnable probability distribution, enabling the generated spectral tokens to exhibit an augmented adaptive focusing capacity, further improving overall recognition accuracy.
	
	The evaluation conclusions manifested in Table \ref{table_4_3_2} are consistent with the aforementioned analysis. A primary difference lies in the relatively constrained performance gain observed in (b). This is mainly due to the strong representation inertia induced by the optical prior in DINOv2. Conventional global alignment techniques only accomplish coarse-grained convergence of distributions and struggle to effectively break this inherent structural pattern, limiting the effectiveness of domain transfer. Nevertheless, the proposed method enables fine-grained modeling of modality discrepancies, thereby achieving deep feature alignment and yielding remarkable performance gains.

	\subsection{Parameter Tuning}
	\subsubsection{Loss Function Weight Coefficient $\lambda$ in MCPT}
	The weight coefficient $\lambda$ is introduced to balance the cross-modal symmetric loss and the unsupervised contrastive loss, where the former promotes alignment between different modalities in the feature space, and the latter encourages the model to learn general visual representations from the data. To evaluate the sensitivity to this hyperparameter, experiments with different values of $\lambda$ are conducted on the MSTAR dataset with DINOv2, and the results are shown in Fig. \ref{fig_4_4_1} (a). The results demonstrate that the best performance is achieved at $\lambda = 0.5$, and the framework maintains stable and high accuracy across a wide range of $\lambda$ values. Concretely, the difference between the maximum and minimum accuracies is only $0.97\%$. This marginal variance reflects the robustness of the proposed method with respect to $\lambda$, which stems from the consistent optimization objectives of the two loss functions, as both aim to reduce modality gaps and suppress background noise, facilitating the extraction of discriminative high-dimensional features. Therefore, changing the loss weights does not lead to significant performance degradation.
	
	\begin{figure*}[!t]
		\centering
		\includegraphics[width=\textwidth]{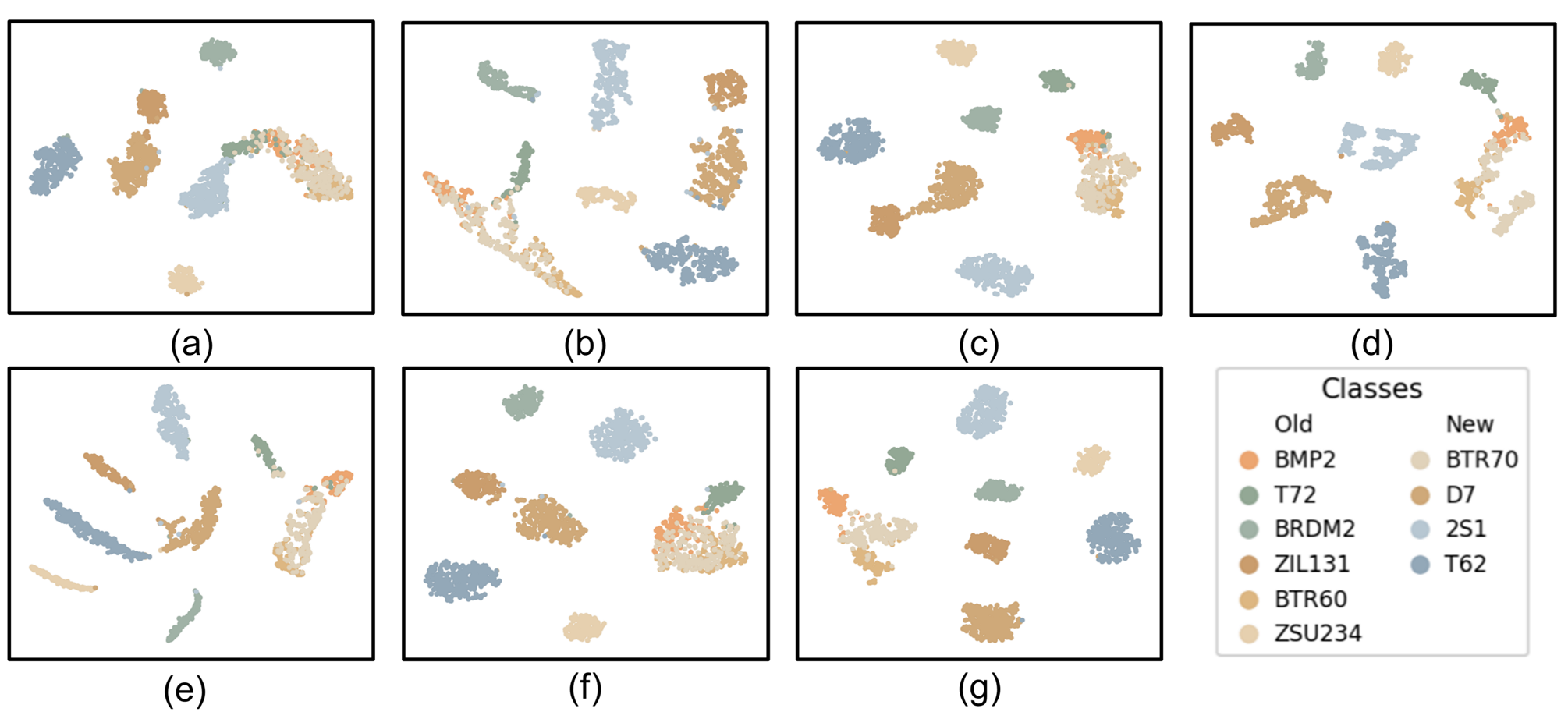}
		\caption{Visualization of features extracted by different methods on the MSTAR dataset with DINOv2. (a) GCD. (b) SimGCD. (c) InfoSieve. (d) CMS. (e) BKD-CL. (f) ProtoGCD. (g) Ours.}
		\label{fig_4_5_1} 
	\end{figure*}

	\subsubsection{Number of Frequency Bands $N$ in the AFT Module}
	The granularity of MDC partitioning in the AFT module is determined by the number of frequency bands $N$, which simultaneously defines the number of experts involved in the subsequent FER-based feature refinement. We increment the value of $N$ from $1$ to $7$ to examine the impact of different partition scales on the efficacy of feature decoupling, with the corresponding performance variations presented in Fig. \ref{fig_4_4_1} (b). Empirical evidence suggests that an optimal trade-off between model size and recognition performance is achieved at $N=4$. Furthermore, recognition accuracy increases as $N$ grows in the initial stage, since finer frequency band division provides more precise feature refinement. Nevertheless, performance exhibits a decline once $N$ surpasses a specific threshold. Such degradation stems from the fact that excessive frequency band segmentation introduces redundancy and coupling among adjacent bands, which weakens the discriminative semantics required for cross-modal alignment. Despite these variations, the evaluation metrics remain stable across the tested range, indicating that the proposed architecture exhibits low sensitivity to the choice of $N$. Notably, when $N=1$, the FER module degenerates into a standard attention-based architecture, where feature interactions rely solely on conventional self-attention mechanisms. Under this setting, the performance gain over the baseline is limited, as frequency-domain decoupling and local refinement are absent.

	\subsubsection{Number of Fine-Tuned Layers $l$ in the Pre-Trained DINO Backbone}
	In the MCPT stage, a partial freezing strategy is applied to the DINO encoder consisting of $12$ Transformer layers, where the shallow layers are kept fixed and only the final $l$ layers are updated. The influence of cross-modal transfer efficacy on the scale of trainable parameters is examined by using representative configurations, ranging from single-layer fine-tuning ($l=1$) and partial fine-tuning ($l=6$) to full-model fine-tuning ($l=12$). The performance histogram shown in Fig. \ref{fig_4_4_1} (c) peaks at $l=2$, followed by a gradual decline as the number of fine-tuned layers increases. This decline is mainly attributed to overfitting caused by optimizing excessive parameters with a constrained set of $2000$ image pairs, which undermines the optical prior learned by DINO from large-scale optical datasets. Moreover, minimum accuracy is observed at $l=1$, since the subsequent SAR-GCD fine-tuning stage also updates only the final layer, resulting in MCPT primarily providing improved parameter initialization instead of enabling the model to learn cross-modal alignment patterns. Consequently, the architecture yields a negligible performance increment of merely $0.73\%$ over the non-pretrained baseline, underscoring the insufficiency of solitary layer updates in capturing complex cross-modal relationships.

	\begin{table}[t]
	\centering
	\caption{Comparison of Intra, Inter and Ratio on the MSTAR Dataset with DINOv2.}
	\begin{tabular}{lllll}
			\toprule
			Backbone & Methods & Intra $\downarrow$ & Inter $\uparrow$ & Ratio $\uparrow$ \\
			\midrule
			\multirow{7}{*}{DINOv2} 
			& GCD       & 7.56  & 59.58          & 7.88  \\
			& SimGCD    & 8.46  & 55.17          & 6.52  \\
			& InfoSieve & 6.41  & \textbf{61.22} & 9.55  \\
			& CMS       & 7.20  & 60.93          & 8.46  \\
			& BKD-CL    & 8.82  & 60.10          & 6.81  \\
			& ProtoGCD  & 6.58  & 59.02          & 8.96  \\
			\cmidrule(lr){2-5}
			& Ours      & \textbf{5.54}  & 57.71 & \textbf{10.41} \\
			\bottomrule
		\end{tabular}%
		\label{table_4_5_1}%
	\end{table}%
	\subsection{Visualization and Analysis}
	\subsubsection{Visualizations of the Feature Space}
	Fig. \ref{fig_4_5_1} illustrates the comparative visualization of features extracted by different algorithms under the DINOv2 architecture. Although current approaches generally accomplish adequate feature clustering on the MSTAR dataset, conventional methods (GCD and SimGCD) struggle to separate the old classes T72 and BMP2 from the novel class BTR70. Besides, significant feature overlap between the old class BMP2 and the novel category BTR70 persists in the feature spaces produced by state-of-the-art models (InfoSieve, CMS, BKD-CL, and ProtoGCD). Conversely, the proposed framework effectively separates the representations of BMP2 and BTR70, achieving effective grouping across all categories and demonstrating strong representation capability.
	
	Moreover, to quantitatively evaluate differences in feature distributions among the algorithms, we introduce three metrics: intra-class variance (Intra), inter-class variance (Inter), and the inter-to-intra-class ratio (Ratio). Table \ref{table_4_5_1} presents the results, where $\downarrow$ denotes that lower values are better, and $\uparrow$ denotes that higher values are better. Consistent with the visual observations, the quantitative results show that the proposed method achieves the best performance in terms of Intra and Ratio. Despite the slightly lower inter-class variance compared to competing methods, this reduction is mainly due to the pronounced clustering effect of the proposed framework, which restricts broader separation among different classes in the learned representations. Interpreting these metrics alongside the visual evidence reveals that the exceptional Intra compactness prevents relatively small inter-class distances from degrading classification performance, ultimately leading to favorable class separability.
	\begin{figure*}[!t]
		\centering
		\includegraphics[width=\textwidth]{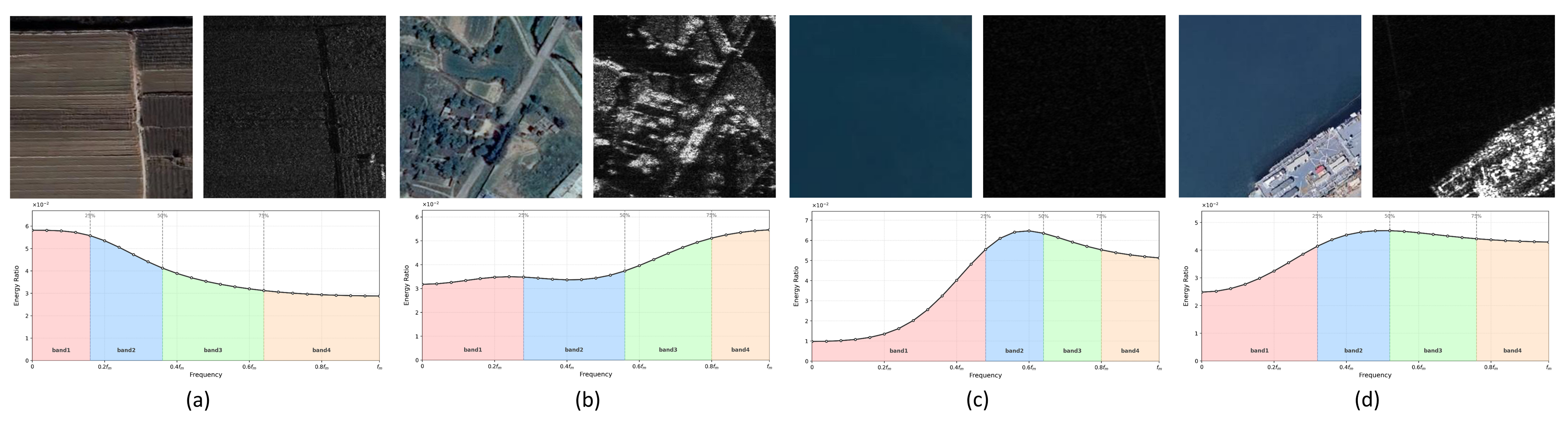}
		\caption{Visualization of MDC for different scenes in the YESeg-OPT-SAR dataset. (a) Farmland. (b) Vegetation. (c) Water. (d) Ship.}
		\label{fig_4_5_2} 
	\end{figure*}

	\begin{figure}[!t]
		\centering
		\includegraphics[width=1\linewidth]{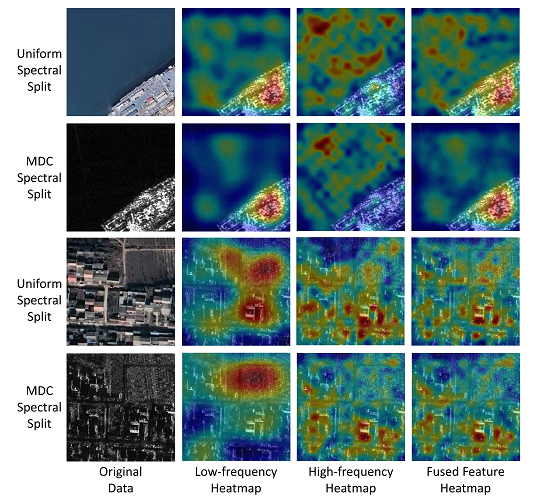}
		\caption{Visualization of frequency token heatmaps in ship and building scenes.}
		\label{fig_4_5_3}
	\end{figure}	
	
	\subsubsection{Visualizations of the MDC}
	Visualization of the MDC provides qualitative evidence that the descriptor captures structured cross-modal discrepancy patterns across different scenes. Fig. \ref{fig_4_5_2} displays these results. The generated curves exhibit clear discriminability across different scenarios while maintaining consistent trends in similar environments. Regarding scenes characterized by significant structural variations, such as farmland, vegetation, and water, the energy distribution curves differ significantly, providing superior discriminative capability. Highly related scenes, including water and ships, produce curves with similar overall patterns. Due to the presence of speckle noise in water regions, both types of curves exhibit manifest elevated energy in the high-frequency spectrum. Building on this foundation, the distinct contour structures of ships generate higher energy responses in the low-frequency band compared to the water scene. These observations confirm that the proposed descriptor, MDC, effectively captures modality differences across frequency components in complex scenarios, thereby providing reliable prior information for subsequent frequency-domain feature refinement.
	
	\subsubsection{Visualization of Frequency Decoupling and Feature Refinement}
	We visualize the frequency tokens generated by the AFT module together with their fused representations in Fig. \ref{fig_4_5_3} to reveal the model’s attention allocation across different frequency bands for cross-modal alignment, thereby supporting that MDC-guided tokenization improves band separation and refinement guidance. Fig. \ref{fig_4_5_3} presents a comparison of frequency-domain representations obtained from the uniform division strategy and the MDC, denoted as UD-Token and MDC-Token, respectively. The ship scenario exhibits significant overlap in the attention distribution of the UD-Token across both low-frequency and high-frequency regions, which indicates limited capability for frequency-band separation. The MDC-Token, however, effectively separates attention between the low-frequency structural contours of the ships and the high-frequency speckle noise, resulting in a more structured frequency-domain representation. For fused features, the proposed MDC-Token focuses alignment attention on ship contours and regions with strong noise responses, whereas the UD-Token assigns redundant attention to large background areas, rendering it less effective in guiding cross-modal alignment. In the building scenario, comparable results are observed, where the MDC-Token suppresses irrelevant background information and concentrates attention on the main building regions. This targeted focus outperforms the UD-Token and highlights the effectiveness of the MDC-Token in frequency-disentangled feature refinement.
	
	\begin{table}[!t]
		\centering
		\caption{Experimental Results of Computational Complexity.}
		\setlength{\tabcolsep}{5pt} 
		\begin{tabular}{llllll}
			\toprule
			\multirow{2}{*}{Method} & \multirow{2}{*}{Stage} & FLOPs    & Params & Trainable  & Size \\
			&       & (GFLOPS) & (M)    & Params (M) & (MB) \\
			\midrule
			Ours     & Pretrain  & 51.73 & 113.39 & 40.99 & 432.56 \\
			(DINOv2)  & Inference & 23.19 & 92.88  & 20.48 & 354.32 \\
			\bottomrule
		\end{tabular}%
		\label{table_4_6_1}%
	\end{table}%
	
	\subsection{Analysis of Computational Complexity}
	To assess the computational cost of our method, we present model complexity during both the pre-training and inference stages in Table \ref{table_4_6_1}. All experiments are conducted using an input resolution of $224 \times 224$ on a single RTX 4090 GPU. In Table \ref{table_4_6_1}, FLOPs denotes floating-point operations, Params the total number of model parameters, Trainable Params the number of parameters updated during training, and Size the model storage in megabytes. The corresponding units are given in parentheses, and lower values are preferable for all metrics. Although the auxiliary AFT and FER modules introduce additional computational overhead during pre-training, this complexity is restricted to the training phase. For inference, these modules are removed, and the model adopts the standard DINOv2 backbone for downstream GCD tasks, so the additional complexity is confined to pretraining only. Consequently, compared to the pre-training stage, the inference process requires approximately 55\% lower FLOPs, 18\% fewer total parameters, 50\% fewer trainable parameters, and an 18\% smaller overall model size.
	
	\section{Conclusion}
	In this paper, we propose MCPT, an optical prior transfer framework that addresses the domain incompatibility of optical-based methods in SAR-GCD by shifting from black-box cross-modal alignment to spectral modeling. The foundation of MCPT lies in the MDC, which characterizes the modality gap between optical and SAR imagery via structured spectral energy, providing spectral guidance. By leveraging this spectral guidance to refine SAR features and facilitate alignment with optical prior, MCPT enables the efficient transfer of optical prior and yields discriminative feature representations for downstream SAR-GCD tasks. Comprehensive experiments on various SAR-ATR datasets demonstrate that the proposed method consistently outperforms state-of-the-art approaches in recognition accuracy. Future work will explore dynamic parameter optimization strategies, such as LoRA, to replace the current fine-tuning of the last two layers during pre-training, thereby enabling more efficient optical-to-SAR transfer in limited paired datasets.

	\bibliographystyle{IEEEtran}
	\bibliography{reference}

	\newcommand{\biovspace}{\vspace{-5cm}}
	
	\begin{IEEEbiography}
		[{\vspace{-0.2in}\includegraphics[width=1in,height=1.10in,clip,keepaspectratio]{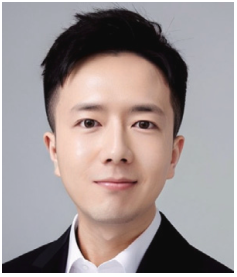}}]
		{Jingyuan Xia} (Member, IEEE) received the B.Sc. and M.Sc. degrees in electrical and electronic engineering from the National University of Defense Technology (NUDT), Changsha, China, in 2014 and 2016, respectively, and the Ph.D. degree in electrical and electronic engineering from Imperial College London (IC), London, U.K., in 2020. His current research interests include machine learning, nonconvex optimization, AIGC, low-level vision, and semantic learning.
	\end{IEEEbiography}
	
	\biovspace	
	
	\begin{IEEEbiography}
		[{\vspace{-0.2in}\includegraphics[width=1in,height=1.10in,clip,keepaspectratio]{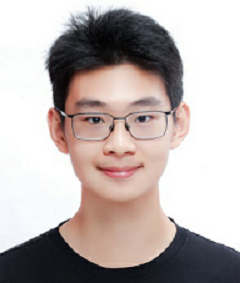}}]
		{Ruikang Hu} received the B.Sc. degree in electronic information engineering from the Guilin University of Electronic Technology (GUET), Guilin, China, in 2024. He is currently working toward the M.Sc. degree in information and communication engineering with the College of Electronic Science, the National University of Defense Technology (NUDT), Changsha, China. His research focuses on deep learning on SAR target recognition.
	\end{IEEEbiography}
	
	\biovspace	
	
	\begin{IEEEbiography}
		[{\vspace{-0.2in}\includegraphics[width=1in,height=1.10in,clip,keepaspectratio]{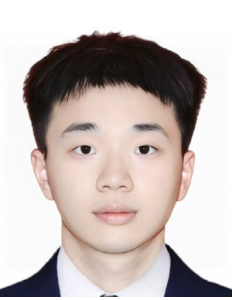}}]
		{Ye Li} received the B.Sc. degree in information engineering from the National University of Defense Technology (NUDT), Changsha, China, in 2023, where he is currently working toward the Ph.D. degree in information and communication engineering with the College of Electronic Science, NUDT. His research focuses on deep learning on long-tailed study.
	\end{IEEEbiography}

	
	\biovspace	
	
	\begin{IEEEbiography}
		[{\vspace{-0.2in}\includegraphics[width=1in,height=1.10in,clip,keepaspectratio]{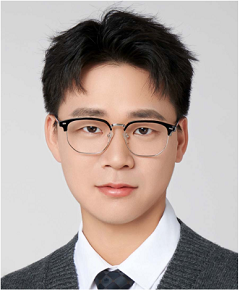}}]
		{Zhixiong Yang} (Student Member, IEEE) received the B.Sc. degree in electronic information engineering from the Northeastern University (NEU), Shenyang, China, in 2021. He is currently working toward the Ph.D. degree in information and communication engineering with the College of Electronic Science, National University of Defense Technology (NUDT), Changsha, China. His research interests include unsupervised learning on image restoration, signal processing, and ISAR imaging.
	\end{IEEEbiography}
	
	\biovspace	
	
	\begin{IEEEbiography}
		[{\vspace{-0.2in}\includegraphics[width=1in,height=1.10in,clip,keepaspectratio]{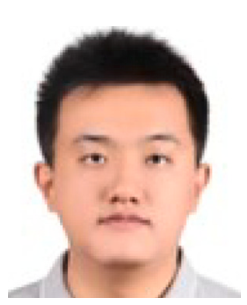}}]
		{Xu Lan} received the B.Sc. degree in information and computing science from the University of Science and Technology Beijing (USTB), Beijing, China, in 2014, the M.Sc. degree in computer science and technology from the National University of Defense Technology (NUDT), Changsha, China, in 2016, and the Ph.D. degree in artificial intelligence from the Queen Mary University of London (QMUL), London, U.K., in 2020. His current research interests include machine learning and large language models.
	\end{IEEEbiography}
	
	\biovspace	
	
	\begin{IEEEbiography}
		[{\vspace{-0.2in}\includegraphics[width=1in,height=1.10in,clip,keepaspectratio]{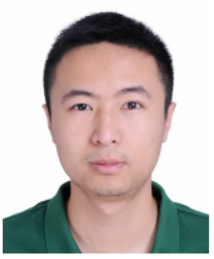}}]
		{Zhejun Lu} received the B.S. degree in communication engineering, and the M.S. and Ph.D. degrees in information and communication engineering from the National University of Defense Technology (NUDT), Changsha, China, in 2011, 2013, and 2017, respectively. His current research interests include target tracking, information fusion and machine learning.
	\end{IEEEbiography}
	
	

\end{document}